\documentclass[letterpaper]{article} 
\usepackage{aaai25}  
\usepackage{times}  
\usepackage{helvet}  
\usepackage{courier}  
\usepackage[hyphens]{url}  
\usepackage{graphicx} 
\usepackage{natbib}  
\usepackage{caption} 
\usepackage{algorithm}
\usepackage{algorithmic}

\usepackage{amsmath}
\usepackage{booktabs}
\usepackage{siunitx}
\usepackage{multirow}
\usepackage{soul}

\usepackage{subfigure}

\usepackage{newfloat}
\usepackage{listings}
\DeclareCaptionStyle{ruled}{labelfont=normalfont,labelsep=colon,strut=off} 
\floatstyle{ruled}
\newfloat{listing}{tb}{lst}{}
\floatname{listing}{Listing}
\title{Reinforcement Learning-Based Prompt Template Stealing for\\ Text-to-Image Models}
\author{
    Xiaotian Zou
}
\affiliations{

    Department of Computer Science, University of Exeter\\ xz549@exeter.ac.uk\\
}

\usepackage{bibentry}

\begin{document}

\maketitle

\begin{abstract}
Multimodal Large Language Models (MLLMs) have transformed text-to-image workflows, allowing designers to create novel visual concepts with unprecedented speed. This progress has given rise to a thriving “prompt-trading” market, where curated prompts that induce trademark styles are bought and sold. Although commercially attractive, prompt trading also introduces a largely unexamined security risk: the prompts themselves can be stolen.

In this paper, we expose this vulnerability and present RLStealer, a reinforcement-learning–based prompt-inversion framework that recovers its template from only a small set of example images. RLStealer treats template stealing as a sequential decision-making problem and employs multiple similarity-based feedback signals as reward functions to effectively explore the prompt space. Comprehensive experiments on publicly available benchmarks demonstrate that RLStealer gets state-of-the-art performance while reducing the total attack cost to under 13\% of that required by existing baselines. Our further analysis confirms that RLStealer can effectively generalize across different image styles to efficiently steal unseen prompt templates. Our study highlights an urgent security threat inherent in prompt trading and lays the groundwork for developing protective standards in the emerging MLLMs marketplace.
\end{abstract}

%

\section{Introduction}
In recent years, the emergence of advanced MLLMs \cite{GPT4V,deepseekvl} and diffusion models \cite{diffusion, dalle3}, have significantly enhanced the performance of text-to-image generation tasks. However, crafting prompts capable of generating high-quality and targeted images typically requires extensive domain expertise along with meticulous fine-tuning and optimization \cite{automaticprompt1,automaticprompt2}, rendering this process highly complex and specialized. This complexity has given rise to a novel business model known as prompt trading.

\begin{figure}[!h]
    \centering
    \includegraphics[width=\columnwidth]{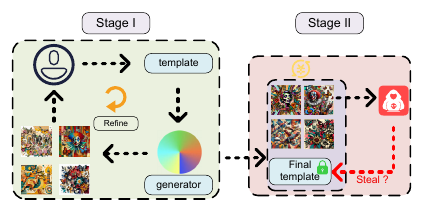}
    \caption{The example of prompt trading.}
    \label{example}
\end{figure}

On dedicated prompt trading platforms such as PromptBase\footnote{https://promptbase.com/} and LaPrompt\footnote{https://laprompt.com/}, creators upload and sell their carefully designed prompts, often accompanied by sample images that demonstrate their unique artistic styles. Upon acquiring these prompts, buyers can modify the subject content while preserving the original artistic style, utilizing corresponding text-to-image generation models to create new images consistent in aesthetics. However, if attackers can infer the templates from merely a few publicly available sample images, it would constitute a severe infringement on creators' intellectual property and threaten the commercial sustainability of prompt trading platforms. Such attacks have been previously identified in literature as prompt template stealing attacks \cite{vulnerability}. For clarity, we illustrate this concept with an example shown in Figure \ref{example}.

Existing studies on prompt stealing fall into two main categories. The first targets a single generated image and attempts to recover the exact prompt that produced it \cite{steal1,steal2,steal3}. The second attacks a set of stylistically consistent images, aiming to infer a general prompt template capable of reproducing the overall style \cite{vulnerability}. Single-image approaches transfer poorly and thus have limited value in real-world settings. Research on template attacks is still nascent—so far only \citet{vulnerability} has addressed the problem, using an evolutionary algorithm. Evolutionary methods, however, are computationally expensive, converge slowly, and struggle to deliver stable performance across diverse scenarios \cite{EA,EA2}, which restricts their practical applicability.

To overcome these limitations, we propose \textbf{RLStealer}, a reinforcement-learning framework for prompt-template stealing. Building on the prompt structure described in \cite{vulnerability,design1}, we decompose a template into three core components—\textbf{Subject}, \textbf{Modifiers}, \textbf{Supplement}—and we construct a discrete action space from a prompt-engineering perspective. We then employ \textit{Proximal Policy Optimization} (PPO) \cite{ppo} as the learning algorithm and design a task-specific reward function, enabling efficient and stable policy optimisation.

Given that only \citet{vulnerability} have studied this problem and introduced the PRISM dataset, we conducted comprehensive experiments on the PRISM dataset to evaluate RLStealer. The results demonstrate that RLStealer not only achieves high-quality prompt template stealing and faithful reconstruction of the original image's artistic style, but also exhibits a low-cost advantage. Our main contributions are as follows:
\begin{itemize}
    \item For the prompt template stealing problem, we innovatively introduce a reinforcement learning approach, designing a novel action space and a customized reward function.
    \item Compared to existing methods, RLStealer significantly reduces the attack cost by at least 87\% while getting state-of-the-art attack performance, demonstrating exceptional efficiency. 
    \item Our study uncovers critical security vulnerabilities in the emerging MLLMs prompt-trading market and establishes a rigorous baseline that lays the groundwork for future security research. 
\end{itemize}

\section{Related Work}
In this section, we review related work from two perspectives: prompt stealing attacks in text-to-image generation and advancements in reinforcement learning.

\subsection{Prompt Stealing Attacks in Text-to-Image Generation}
Prompt stealing attacks, also known as prompt inversion attacks, involve inferring the original input prompt by analyzing the output content generated by a model \cite{steal1,steal2,steal3,vulnerability}. Successful execution of such attacks not only risks infringing on creators’ intellectual property but also poses a significant threat to the business models and trust foundations of prompt trading platforms. Compared to text generation tasks, the inherent randomness and semantic ambiguity in image generation processes make such attacks particularly complex and challenging. Current text-to-image prompt stealing attacks can be broadly categorized into two types: 

\textbf{(a) Prompt Stealing for Single Images: These methods aim to recover the exact prompt used to generate a specific image.}
\begin{itemize}
    \item CLIP Interrogator leverages the CLIP model \cite{clip} to identify core content in images and selects the most matching descriptions from a predefined phrase library \cite{steal1}.
    \item \citet{steal2} proposed a method that fine-tunes two separate models to extract the subject elements and stylistic features of an image, subsequently integrating these to construct a complete prompt.
    \item \citet{steal3} further optimized this approach by employing GPT-4V for multi-round iterative refinement of prompts, resulting in richer and more consistent prompt texts.
\end{itemize}

\textbf{(b) General Prompt Template Stealing for Multiple Images: These methods focus on extracting generalizable prompt templates from a set of stylistically similar images.}
\begin{itemize}
    \item EvoStealer \cite{vulnerability} adopts a differential evolutionary algorithm to steal universal templates capable of generating a series of images with a specific style.
\end{itemize}

However, both categories of methods exhibit notable limitations. Single-image prompt acquisition methods lack generalization, as they are tailored to recovering prompts for specific images and struggle to meet diverse real-world application needs. Conversely, template stealing methods, such as those based on evolutionary algorithms, often require extensive model queries (resulting in high interaction costs) or substantial computational resources, making it challenging to achieve an optimal balance between attack cost and effectiveness.

\subsection{Advancements in Reinforcement Learning}
Since \citet{RL} systematically introduced the reinforcement learning (RL) framework, the field has witnessed remarkable progress. The Deep Q-Network (DQN) marked a milestone by successfully integrating deep neural networks with RL, ushering in the era of deep reinforcement learning \cite{DQL1,DQL2}. Subsequently, PPO algorithm \cite{ppo}, based on the actor-critic framework, has been widely recognized as one of the most advanced RL methods due to its stability and efficiency. PPO has played a pivotal role in reinforcement learning with human feedback \cite{RLHF}. Notably, the DeepSeek-R1 \cite{deepseekr1} proposed by DeepSeekAI in 2025 demonstrated that training LLMs primarily through RL can significantly enhance their reasoning capabilities. This breakthrough underscores the immense value and broad potential of RL in unlocking the capabilities of large models and modeling intelligent behaviors.

\section{Problem Formulation}
In a typical prompt trading platform, creators upload a core prompt template \( T \) (visible to consumers only after purchase). To demonstrate the artistic style and effectiveness of the template, creators preset or substitute specific "subject" placeholders (or content cores) within \( T \), generating multiple derived prompts. These prompts are then input into a designated text-to-image generation model \textbf{V} (e.g., DALL·E 3) to produce a set of example images \( S = \{s_1, s_2, \ldots, s_N\} \), where \( N \) represents the total number of example images for template \( T \). These images share the same artistic style but vary in subject content. After purchasing \( T \), consumers can similarly replace the subject content and use the same \textbf{V} model to generate customized images with a consistent style.

\paragraph{Attacker’s knowledge}
In this study, we define the attacker's environment as follows:
\begin{itemize}
    \item \underline{Observable Information}: The attacker has access only to the publicly showcased example image set \( S \) and the model information of \textbf{V} used to generate these images. The original prompt template \( T \) remains unknown to the attacker.
    \item \underline{Auxiliary Resources}: The attacker can access a collection of publicly available, free prompt templates \( F = \{f_1, f_2, \ldots, f_q\} \), where \( q \) is the total number of free templates. For each free template \( f_k \in F \), the attacker also has access to its corresponding example image set \( W_k = \{w_{k,1}, w_{k,2}, \ldots, w_{k,m}\} \), where \( m \) is the number of example images for template \( f_k \), generated by model \textbf{V}.
\end{itemize}

\paragraph{Attacker’s objective}

Given the above information, the attacker seeks an estimated template
\(\hat{T}\) such that  
(i) images rendered from \(\hat{T}\) resemble the exemplar set \(S\) as closely
as possible, and  
(ii) the template text \(\hat{T}\) itself is as similar as possible to the
hidden ground-truth template \(T\).

These assumptions align with the operational model and information disclosure practices of mainstream prompt trading platforms such as \textbf{PromptBase}.

\section{RLStealer}
\begin{figure*}[!h]
    \centering
    \includegraphics[width=\textwidth]{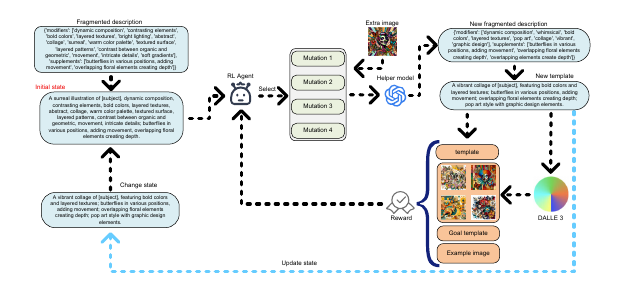}
    \caption{The framework of RLStealer. In this framework, given a set of example images generated from the same prompt template, we first perform a warm start to obtain an fragmented description and initial template. This initial state is then fed into the RL agent, which selects an appropriate mutation action. A helper model executes the selected action to generate a new descriptive state. This description is combined with a subject to construct a complete prompt, which is used to generate an image. The generated image, together with the textual description, the target template, and reference images, is then used to compute a reward signal that updates the RL agent.}
    \label{framework}
\end{figure*}

In this section, we first present an overview of our system. Next, because we cast the prompt‑template stealing problem as a reinforcement‑learning task, we introduce three task-specific extensions: 
(\emph{i}) a \emph{warm-start} procedure that supplies the policy with an informative initial state; 
(\emph{ii}) a task-aligned, discrete \emph{action space} that directly edits the \textbf{Modifiers}, \textbf{Supplement} fields; and (\emph{iii}) a \emph{multi-component reward} that balances textual fidelity with visual consistency. For ease of understanding, the framework of RLStealer is illustrated in Figure \ref{framework}.

\subsection{Overview of RLStealer}
\label{sec:tech_details}

\paragraph{RL formulation.}
We cast our system as a Markov decision process
$\mathcal{M}=\langle\mathcal{S},\mathcal{A},\mathcal{T},\mathcal{R},\gamma\rangle$
in which  
$\mathcal{S}$ is the state space,  
$\mathcal{A}$ the discrete action space,  
$\mathcal{T}:\mathcal{S}\!\times\!\mathcal{A}\!\to\!\mathcal{S}$ the state‑transition dynamics,  
$\mathcal{R}:\mathcal{S}\!\times\!\mathcal{A}\!\to\!{R}$ the multi‑component reward of Eq. (\ref{reward_equ}),  
and $\gamma\!\in\!(0,1)$ a discount factor.
The agent seeks a policy $\pi_\theta$ that maximizes the expected return
$\mathbf{E}_{\pi_\theta}\!\left[\sum_{t=0}^{T}\gamma^{t}r_t\right]$. The transition function $\mathcal{T}$ is not accessible so we could rely on model‑free learning algorithm.

\paragraph{State representation.}
At time step $t$ the environment produces a fragmented description. Then we feed the fragmented description into an LLM and ask the model to transform into a new template $p^{(t+1)}$. The prompt is shown in Supplementary Materials. Finally, We obtain the hidden representation of $p^{(t+1)}$ using the RoBERTa family of models \cite{robert,robert2} and treat it as $s^{(t+1)}$.

\paragraph{Action and Reward.}
We designed four actions and a reward function composed of three components.

\paragraph{Optimisation algorithm.}
We adopt PPO \cite{ppo}, which enjoys strong empirical
stability. Given trajectories collected under the old policy $\pi_{\theta_{\mathrm{old}}}$, PPO maximizes
the clipped surrogate objective
\begin{equation}
\label{eq:ppo}
\begin{split}
\mathcal{L}_{\text{PPO}}(\theta)=
{E}_{(s^{(t)},a^{(t)})\sim\pi_{\theta_{\mathrm{old}}}}
\Bigl[
\min\!\bigl(
r_\theta^{(t)}A^{(t)}, \\
\operatorname{clip}(r_\theta^{(t)},1-\varepsilon,1+\varepsilon)A^{(t)}
\bigr)
\Bigr],
\end{split}
\end{equation}
where
$r_\theta^{(t)}\!=\frac{\!\pi_\theta(a^{(t)}| s^{(t)})}{\pi_{\theta_{\mathrm{old}}}(a^{(t)}|s^{(t)})}$
and $A^{(t)}=R^{(t)}-V^{t}$, $R^{(t)}=\sum_{k=t+1}^{T}\gamma^{\,k-t-1}\,r^{(k)}$.

\subsection{Warm Start}
The quality of the initial state critically affects subsequent learning efficiency.  
RLStealer begins by randomly sampling \(n\) images from the \(W_k\) exemplars linked to template \(f_k\) to obtain the collection of initial descriptions. Following the preprocessing strategy of EvoSteal, each image is analysed with GPT-4o to produce a structured \emph{fragmented description}
$\langle\textbf{Subject}, \textbf{Modifiers}, \textbf{Supplement}\,\rangle$.
This step yields the collection \(\{e_1,e_2,\dots,e_n\}\).

Unlike population-based approaches, where every individual in the population can serve as an initial state, RLStealer can begin only from the fragmented description of a single state. Consequently, we aggregate the fragments \(\{e_1,e_2,\dots,e_n\}\) and invoke GPT-4o a second time to synthesise their shared stylistic attributes into one \emph{Initial Summarised Description}. This compact summary—representing the overall style of the target template—serves as the starting state for the RLStealer.

\subsection{Action Space Design}
\label{sec:action}
Given our goal of extracting generalizable \textbf{Modifiers} and \textbf{Supplement} as the core components of a prompt template (with the Subject being variable), we design the action to focus on optimizing these two components (The prompt for each action is shown in Supplementary Materials. Specifically, we define the following four types of action, which the agent can select to progressively refine the current fragmented description:
\begin{itemize}
    \item \ul{Action 1 (Preserve Commonality - Deterministic Combination)}: For two input fragmented descriptions, extract and retain the common elements of their "Modifiers" and "Supplements", combining them to form a new fragmented description.
    \item \ul{Action 2 (Preserve Commonality - Random Combination)}: For two input fragmented descriptions, extract and retain the common elements of their "Modifiers", then randomly select one (or a combination, with a specified strategy) of the "Supplements" from the two inputs to generate a new fragmented description.
    \item \ul{Action 3 (Differential Mutation)}: For two input fragmented descriptions, identify differences in their "Modifiers" and "Supplements". Then, guided by the original example images (e.g., via visual features or keywords extracted from the images) and the characteristics of the current fragmented description, perform targeted mutations on the differing parts to explore and generate a new fragmented description.
    \item \ul{Action 4 (Image-Guided Cross-Fusion)}: For two input fragmented descriptions, randomly select one original example image. Through a mechanism (e.g., using image information to guide the fusion of the two textual description fragments or incorporating key image information into the description), perform a cross-operation between the image information and the two fragmented descriptions to generate a new fragmented description.
\end{itemize}

\subsection{Reward Function Design}
To guide the agent toward learning an effective template-stealing strategy, we design a multi-component reward function. After the agent executes an action and generates a new fragmented description, we first use a LLM to normalize it and convert it into an executable template description \( t \) (with the \textbf{Subject} as a placeholder). The reward function comprises the following three components:
\begin{itemize}
    \item Reward $1$ (Text-Image Matching Score): Replace the Subject placeholder in template $t$ with the corresponding subject content of each original example image to generate complete prompts$\{t_{k,1},t_{k,2}...t_{k,m}\}$. Then compute the average cosine similarity between the embedding vectors of these prompts and the original example images. This reward is designed to encourage the generated template to closely align with the style and content of the images. The $r_1$ is formulated as:
    \[
    \text{r}_1 = \frac{1}{m}\sum_{i=1}^{m}similarity(t_{k,i},w_{k,i})
    \]
    \item Reward $2$ (Sampled Image Matching Score): To improve computational efficiency, randomly select one of the \( m \) original example images and its corresponding subject. Insert this subject into the Subject placeholder of template \( t \), generate a complete prompt, and input it into \textbf{V} to generate an image $w_{sample}$. Compute the embedding vector similarity between this newly generated image and the selected original example image. The $r_2$ is formulated as:
    \[
    \text{r}_2 = similarity(w_{sample},w_{k,random})
    \]
    \item Reward $3$ (Target Template Approximation): During the training phase, the attacker has access to a publicly available dataset, which includes the target prompt templates. We incorporate this into the reward design. Specifically, we compute the embedding vector similarity between the currently generated template description $t$ and the ground-truth target prompt template $f_k$. This reward directly measures the closeness between the stolen template and the true template. The $r_3$ is formulated as:
    \[
\text{r}_3 = similarity(t,f_k)
\].
\end{itemize}
The final reward function is formulated as:
\begin{equation}
\label{reward_equ}
    \text{r} = \lambda_1 \cdot \text{r}_1 + \lambda_2 \cdot \text{r}_2 + \lambda_3 \cdot \text{r}_3
\end{equation}
where \(\lambda_1\), \(\lambda_2\), and \(\lambda_3\) are weighting coefficients.


\section{Experiments}
In this section, we present a comprehensive overview of our experiments. We begin by introducing the datasets used in our study, followed by a detailed description of the baseline methods for comparison. Subsequently, we then outline the experimental setup, evaluation metrics and analysis of the results. Finally, we conduct a series of ablation studies.

\begin{table*}[ht]
    \centering
    \small
    \setlength\tabcolsep{6.8pt} 
    \renewcommand{\arraystretch}{1}
    \scalebox{1.0}{ 
    \begin{tabular}{c|c|c|c|c|c|c|c}
        \toprule
        \textbf{Method} & \textbf{DINO } & \( \textbf{CLIP}_{\textit{img}} \) & \( \textbf{CLIP}_{\textit{txt}} \) & \( \textbf{SigLIp}_{\textit{img}} \) & \( \textbf{SigLIp}_{\textit{txt}} \) & \textbf{Average} & \textbf{Queries}\\
        \midrule
        \multicolumn{8}{c}{\textit{Easy Benchmark}} \\ 
        \midrule\midrule
        BLIP 2 \cite{steal1}  & 62.07 & 79.38 & 48.35 & 82.32 & 52.69 & 64.96 & 0\\
        \midrule
        CLIP Interrogator \cite{steal2} & 69.93 & 82.76 & 54.14 & 85.86 & 62.59 & 70.86 & 0\\
        \midrule
        PromptStealer \cite{steal3}  & 63.73 & 77.90 & 49.21 & 82.73 & 61.93 & 67.10 & 0\\
        \midrule
        EvoStealer \cite{vulnerability} & \textbf{75.83} & \textbf{85.30} & \underline{74.41} & \textbf{89.14} & \underline{72.25} & \underline{79.49} & 25\\
        \midrule
        \textbf{RLStealer (ours)}   & \underline{73.87}  & \underline{85.11} & \textbf{79.98} & \underline{86.89} & \textbf{74.63} & \textbf{80.10} & 0\\

        \midrule
        \multicolumn{8}{c}{\textit{Hard Benchmark}} \\ 
        \midrule\midrule
        BLIP 2 \cite{steal1} &61.16 & 76.67 & 46.04 & 80.51 & 50.74 & 63.02 & 0\\ \midrule
        CLIP Interrogator \cite{steal2} & 66.45 & 78.26 & 54.62 & 82.45 & 60.78 & 68.51 & 0\\ \midrule
        PromptStealer \cite{steal3}  & 60.01 & 75.58 & 47.10 & 79.20 & 59.71 & 64.32 & 0\\ \midrule
        EvoStealer \cite{vulnerability} & \underline{69.24} & \underline{81.34} & \underline{70.61} & \textbf{85.28} & \textbf{69.27} & \underline{75.15} & 25\\
        \midrule
        \textbf{RLStealer (ours)}  & \textbf{70.20} & \textbf{84.00} & \textbf{73.52} & \underline{84.32} & \underline{66.28} & \textbf{75.66} & 0\\                                                          
        \bottomrule
    \end{tabular}}
    \caption{The table presents the overall results on the in-domain data: bold values mark the best scores, while underlined values indicate the second-best. We define the queries as the number of times the attacker needs to query model \textbf{V} in order to obtain the template.
}
    \label{tb:main_indomain}
\end{table*}

\subsection{Dataset}
Given that the PRISM dataset is currently the only available benchmark for this task, we conduct systematic experiments using this dataset. It contains a total of $50$ prompt templates, with $25$ labeled as “Easy” and the remaining $25$ labeled as “Hard” based on their stealing difficulty. Additionally, for each template, the first $5$ images are designated as in-domain data, while the remaining $4$ are considered out-of-domain. In our study, since we need to select a subset of data as publicly accessible to the attacker, we choose $20$ prompt templates for training, including $8$ “Easy” and $12$ “Hard” templates. For each selected template, we use all $5$ corresponding in-domain images as training samples.

\begin{table*}[ht]
    \centering
    \small
    \setlength\tabcolsep{6.8pt} 
    \renewcommand{\arraystretch}{1}
    \scalebox{1.0}{ 
    \begin{tabular}{c|c|c|c|c|c|c|c}
        \toprule
        \textbf{Method} & \textbf{DINO } & \( \textbf{CLIP}_{\textit{img}} \) & \( \textbf{CLIP}_{\textit{txt}} \) & \( \textbf{SigLIp}_{\textit{img}} \) & \( \textbf{SigLIp}_{\textit{txt}} \) & \textbf{Average} & \textbf{Queries}\\
        \midrule
        \multicolumn{8}{c}{\textit{Easy Benchmark}} \\ 
        \midrule\midrule
        CLIP Interrogator \cite{steal2} & 64.02 & 78.72 & 53.95 & 82.98 & 63.73 & 68.68 & 0\\ \midrule
        PromptStealer \cite{steal3} & 60.53 & 75.53 & 51.37 & 81.19 & 61.16 & 65.96 & 0\\ \midrule
        EvoStealer \cite{vulnerability} & \textbf{75.14} & \underline{83.91} & \underline{74.18} & \textbf{85.75} & \underline{73.53} & \underline{79.10} & 25\\ 
        \midrule
        \textbf{RLStealer (ours)}   & \underline{75.12} & \textbf{85.04} & \textbf{79.84} & \underline{85.68} & \textbf{75.14} & \textbf{80.16} & 0\\  
        
        \midrule
        \multicolumn{8}{c}{\textit{Hard Benchmark}} \\ 
        \midrule\midrule
        CLIP Interrogator \cite{steal2} & 62.23 & 69.90 & 51.66 & 75.19 & 58.51 & 63.50 & 0\\ \midrule
        PromptStealer \cite{steal3} & 58.53 & 70.42 & 45.29 & 74.38 & 55.07 & 60.74 & 0\\ \midrule
        EvoStealer \cite{vulnerability} & \textbf{67.00} & \underline{80.50} & \underline{69.27} & \underline{84.55} & \textbf{69.79} & \underline{74.22} & 25\\
        \midrule
        \textbf{RLStealer (ours)}   & \underline{66.68} & \textbf{83.54} & \textbf{73.04} & \textbf{84.59} & \underline{68.38} & \textbf{75.25} & 0\\   
        \bottomrule
    \end{tabular}}
    \caption{The overall evaluation results for the out-domain data.}
    \label{tb:main_outdomain}
\end{table*}

\subsection{Baseline Methods}
Our comparative experiments include state-of-the-art image captioning models and prompt stealing attack methods, specifically:
\begin{itemize}
    \item BLIP-2: BLIP-2 \cite{steal1} is an advanced multimodal pretraining architecture. It employs a lightweight Querying Transformer to effectively bridge a frozen image encoder with LLMs, achieving efficient alignment between images and text. In this study, we use the BLIP-2-opt-2.7b version to generate textual descriptions from images.
    \item CLIP Interrogator: CLIP Interrogator \cite{steal2} is an image content analysis tool based on the CLIP model \cite{clip}. It leverages predefined prompt categories (e.g., artistic style, medium, artist) to assist in generating image descriptions. Its core mechanism involves encoding both images and prompt texts and identifying the textual description most aligned with the given image by computing feature similarity.
    \item PromptStealer: PromptStealer \cite{steal3} is a dedicated attack method for prompt stealing, consisting of two key modules: (1) Subject Generator: Built upon the BLIP model with fine-tuning, this module accurately identifies the core objects in images. (2) Modifier Detector: A multi-class classifier that selects the most style-aligned modifiers from a predefined style vocabulary. PromptStealer combines the identified subject with the detected style modifiers to construct a complete prompt template.
    \item EvoStealer: \citet{vulnerability} employed a differential evolution algorithm for template stealing. In this method, GPT-4o is utilized to assist in processing and optimizing fragmented descriptions (e.g., supporting crossover and mutation operations or refining their results) and ultimately organizes the optimized fragmented descriptions into a structurally complete prompt template.
\end{itemize}

\subsection{Experimental Setup}
In our experiments, the RLStealer agent is implemented as a multilayer perceptron with a $768$-dimensional input, $256$ hidden units, and $4$ layers. The value function is represented by an MLP with a $768$-dimensional input, $256$ hidden units, and a single output layer. We use the SigLIP model to compute text and image embeddings for calculating the reward components \( \text{reward}_1 \), \( \text{reward}_2 \), and \( \text{reward}_3 \). The weights for the reward components are set as \( \lambda_1 = 0.4 \), \( \lambda_2 = 0.4 \), and \( \lambda_3 = 0.2 \). When generating images from the derived prompts, we uniformly use the DALL·E 3 model. The temperature parameter of all relevant language models is set to $0$ to ensure deterministic outputs. During training, the learning rate is set to $0.03$, each episode consists of $8$ steps, and the batch size is $16$. For each batch, we use Adam \cite{adam} to perform $32$ optimization updates.

\subsection{Evaluation Metrics}
We adopt the following widely accepted metrics \cite{steal1,steal2,steal3} to comprehensively evaluate the performance of RLStealer and the baseline methods:
\begin{itemize}
    \item Subject Similarity: To assess the consistency of core subject content between image pairs, we use the self-supervised visual model DINO \cite{dinov2} for feature extraction and comparison. Subject content is a critical dimension for measuring image similarity, making this metric highly relevant.
    \item Style Similarity: To evaluate style consistency, we employ CLIP \cite{clip} and SigLIP \cite{siglip} to extract style features from images. We quantify style similarity by comparing the differences between images generated from stolen prompts and the original reference images in these feature spaces.
    \item Semantic Similarity: To measure the semantic closeness between stolen prompts and the original target prompts, we use CLIP and SigLIP to extract text embeddings for both and compute their cosine similarity.
\end{itemize}

\subsection{Experiments results}
We present the comparison results of RLStealer and other methods on the in-domain data in Table \ref{tb:main_indomain}:
\begin{itemize}
    \item On the easy benchmark, RLStealer achieves an average score of $\textbf{80.10}$, outperforming EvoStealer’s $79.49$. In particular, it sets new highs on both $\textbf{CLIP}_{\textit{txt}}$ ($79.98$ vs.\ $74.41$) and average score, while requiring $0$ query to the target model (EvoStealer uses $25$). This represents a substantial improvement in both attack efficacy and cost. In addition, RLStealer exceeds all other baselines by at least 1 point on average, and by over 5 points on $\textbf{CLIP}_{\textit{txt}}$.  

    \item On the hard benchmark, RLStealer attains an average of $\textbf{75.66}$, again topping EvoStealer’s $75.15$. It leads on \textbf{DINO } ($70.20$ vs.\ $69.24$), $\textbf{CLIP}_\textit{img}$ ($84.00$ vs.\ $81.34$) and $\textbf{CLIP}_\textit{txt}$ ($73.52$ vs.\ $70.61$), demonstrating robust performance even under more challenging templates, all while still issuing $0$ queries.
\end{itemize}

We present the comparison results of RLStealer and other methods on the out-domain data in Table \ref{tb:main_outdomain} \footnote{Due to the inherent challenges associated with accurately identifying and replacing subject entities in prompts generated by BLIP-2, its evaluation is restricted to in-domain data only \cite{vulnerability}.}:
\begin{itemize}
    \item On the easy benchmark, RLStealer’s average of $\textbf{80.16}$ comfortably exceeds EvoStealer’s $79.10$. It also records the best scores on $\textbf{CLIP}_\textit{img}$ ($85.04$ vs.\ $83.91$) and $\textbf{CLIP}_\textit{txt}$ ($79.84$ vs.\ $74.18$), confirming that our method maintains both visual style and semantic fidelity when applied to novel subjects without additional model queries.

    \item On the hard benchmark, with an average of $\textbf{75.25}$, RLStealer again outperforms EvoStealer ($74.22$) and all other baselines. It leads on $\textbf{CLIP}_\textit{img}$ ($83.54$ vs.\ $80.50$) and $\textbf{CLIP}_\textit{txt}$ ($73.04$ vs.\ $69.27$), showcasing strong generalization to unseen content.
\end{itemize}

To give an intuitive visual impression, we provide in Supplementary Materials example images generated by each method under both the easy benchmark and the hard benchmark.

In addition to tracking the number of queries made to model \textbf{V} during the attack, we further analyze the overall attack cost. Since image generation constitutes the primary resource expenditure in our experiments, we use the number of calls to model \textbf{V} as the core metric for cost estimation. During the training phase, we utilize $20$ prompt templates, each generating 8 images, resulting in a total of 160 queries to model \textbf{V}. Based on DALL·E 3’s pricing of \$$0.04$ per image, the total training cost amounts to \$$6.40$. Our experiments involve 50 prompt templates in total, which translates to an average cost of only \$$\textbf{0.128}$ per template. More importantly, the training process is performed only once, and the trained framework can be reused to steal an unlimited number of prompt templates, offering excellent scalability. In contrast, EvoStealer requires $25$ queries to model \textbf{V} for each template, resulting in a per-template cost of approximately \$$\textbf{1.00}$. Therefore, RLStealer achieves comparable or even superior performance to existing state-of-the-art methods while reducing the attack cost to less than $13$\% of EvoStealer, demonstrating a significant advantage in cost-effectiveness.

\begin{table}[!t]
    \centering
    \small
    \setlength\tabcolsep{8.8pt} 
    \renewcommand{\arraystretch}{0.9}  

    \begin{tabular}{c|c|c|c}
        \toprule
        \textbf{Method} & \textbf{InDom.} & \textbf{OutDom.} & \textbf{Average} \\
        \midrule
        \hspace{2pt}Easy   & 78.35 (-) & 77.91 (-) & 78.13 (-) \\ \midrule
        \hspace{2pt}Hard    & 74.77 (-) & 75.11 (-) & 74.94 (-) \\ 
        \bottomrule
    \end{tabular}
    \caption{The results obtained by random strategy.
}
    \label{tb:abalation}
\end{table}

\subsection{Ablation Study: Random Mutation with Fitness Pooling}
To assess the necessity of reinforcement learning in RLStealer, we introduce a random action baseline. In this framework, at each step, a action operator is randomly selected from action 1-4. Critically, after each step, the candidate template is added to a shared pool and assigned a fitness score based on $r_1$ (i.e., average image similarity with reference images). After $8$ random steps, the template with the highest fitness score is selected as the final output for image generation. This process avoids policy learning and instead relies purely on random search with fitness selection.

\begin{figure}[!h]
\centering
\subfigure[The score distribution of each metric for RLStealer and the random strategy under the easy benchmark.]{
\includegraphics[width=0.45\columnwidth]{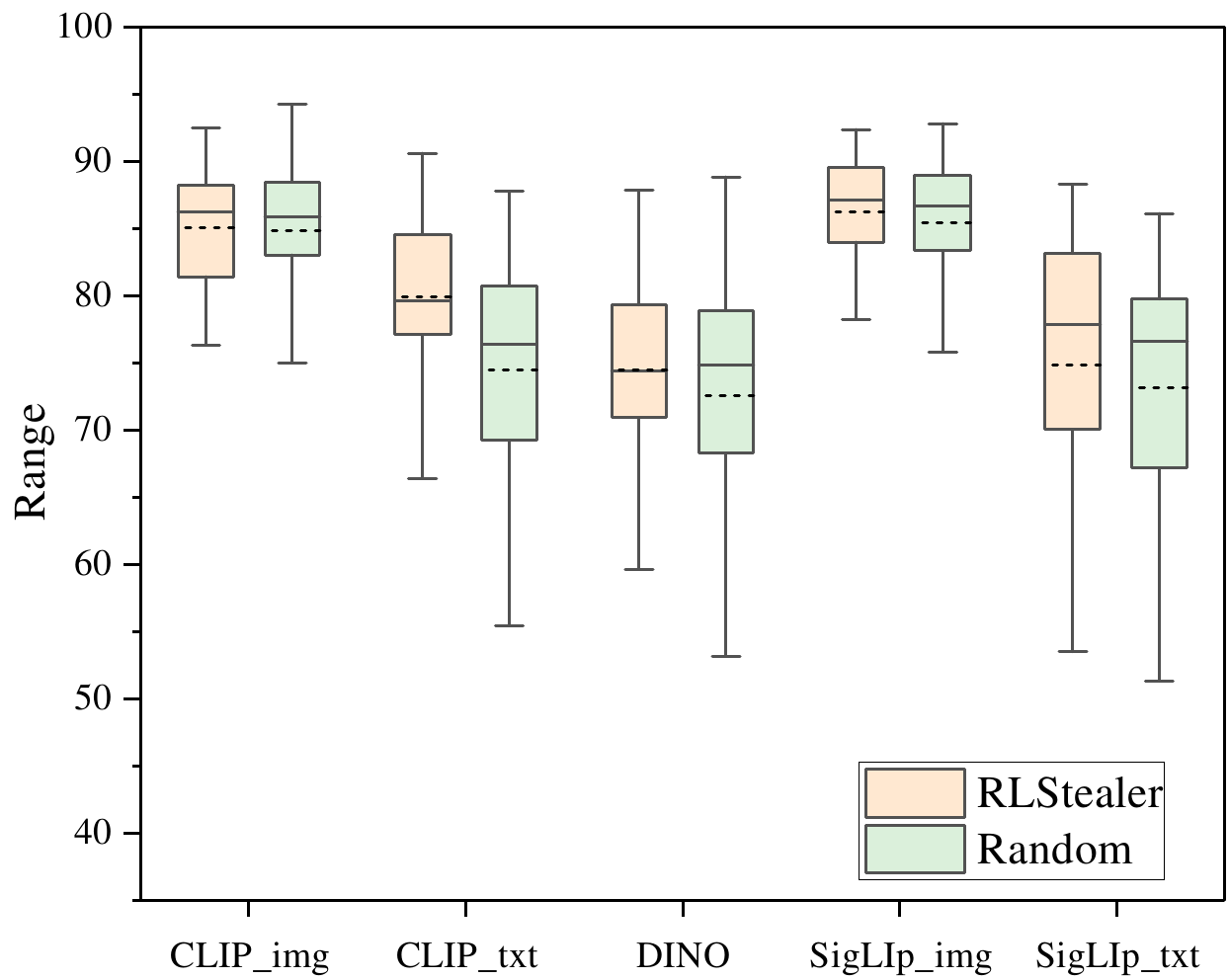}
}
\subfigure[The score distribution of each metric for RLStealer and the random strategy under the hard benchmark.]{
\includegraphics[width=0.45\columnwidth]{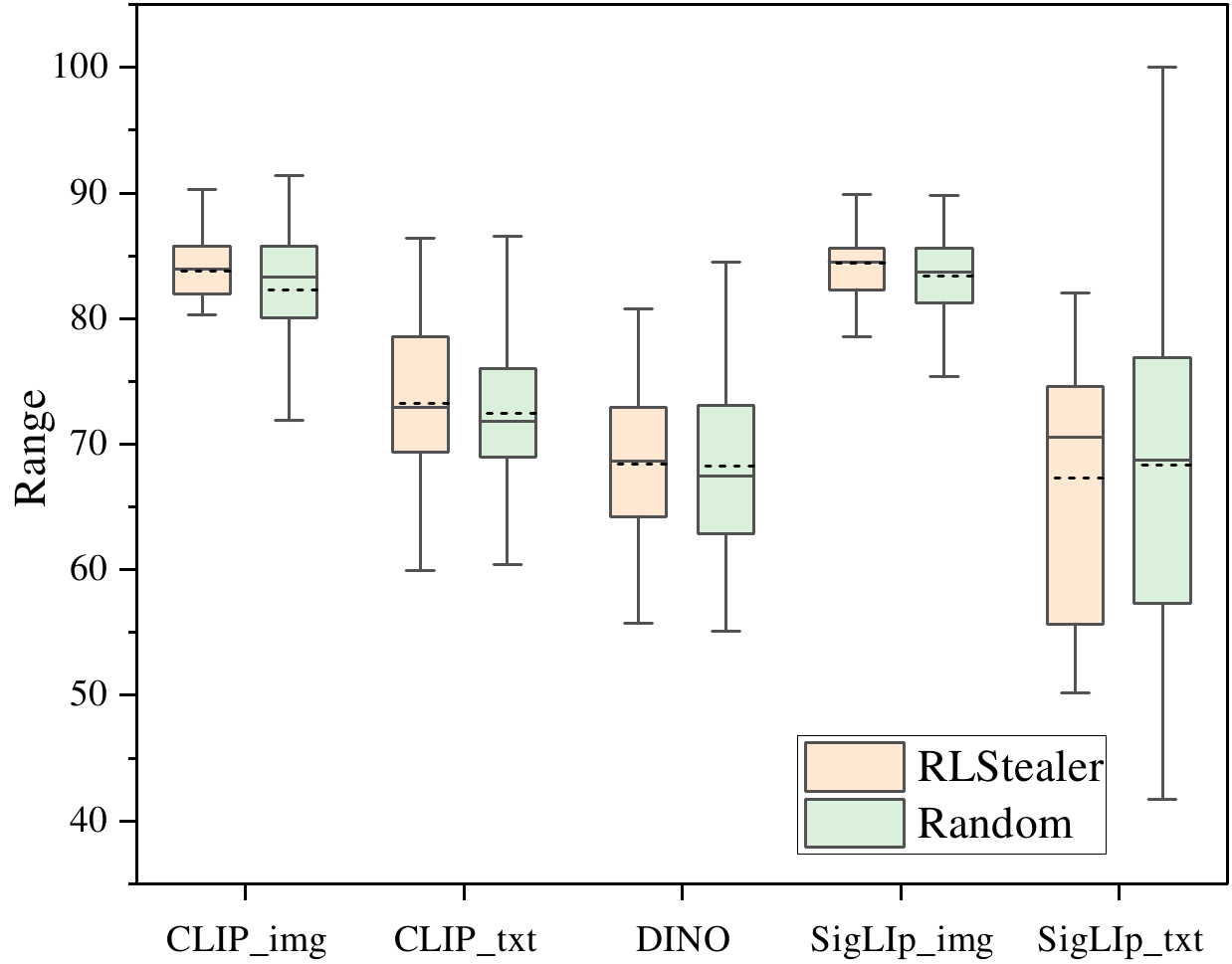}
}
\caption{The score distribution of each metric for RLStealer and the random strategy under the easy and hard benchmarks.}\label{var}
\end{figure}

The results are summarized in Table \ref{tb:abalation}, where the reported InDom. and OutDom. scores represent the average performance across five evaluation metrics: \textbf{DINO }, \( \textbf{CLIP}_{\textit{img}} \), \( \textbf{CLIP}_{\textit{txt}} \), \( \textbf{SigLIp}_{\textit{img}} \), \( \textbf{SigLIp}_{\textit{txt}} \).
\begin{itemize}
    \item On the easy benchmark, the random strategy scores $78.35$ in-domain and $77.91$ out-of-domain, for an overall average of $78.13$—substantially below RLStealer’s $80.10$/$80.16$. This gap indicates that in simple template scenarios, blind action selection and fitness-based filtering lack the guided exploration, making it difficult to consistently discover high-quality templates.

    \item On the hard benchmark, the random strategy scores $74.77$ in-domain and $75.11$ out-of-domain, for an overall average of $74.94$—close to RLStealer’s $75.66$/$75.25$. This shows that for more complex templates, pure random search combined with fitness selection can occasionally stumble on strong solutions.

    \item Furthermore, as shown in Figure \ref{var}, we observe that the random strategy exhibits greater variability under both the easy and hard benchmarks. Although it may occasionally achieve performance close to or even surpassing that of RLStealer in some trials, its large fluctuations introduce a risk of instability. In contrast, RLStealer ensures consistently reliable output quality with a higher median performance and lower volatility.
\end{itemize}


\section{Conclusion and Limitation}
\paragraph{Conclusion.} We presents RLStealer—the first reinforcement-learning framework explicitly designed to steal prompt templates from text-to-image models. We construct a discrete action space and devise a multi-component reward, thereby recasting template extraction as a sequential decision problem solvable with PPO. Extensive experiments demonstrate that RLStealer gets state-of-the-art performance while slashing attack cost to only 13\% of existing methods. The agent also generalises well to unseen subjects, artistic styles, and sampling budgets. An ablation study with random strategy confirms that the performance gains arise from guided policy learning rather than random search. Looking ahead, we will explore defences that mitigate such attacks.  We urge the research community to devote broader attention to this issue and to pave the way for secure prompt trading in MLLMs ecosystems.

\paragraph{Limitation.}
The main limitation of our study is cost. Every query to the commercial text‑to‑image model DALL·E 3 is priced at \$$0.04$. A single training run of RLStealer generates roughly $160$ images, and one full evaluation adds about $500$ more. Each baseline comparison requires at least $450$ additional images. Under these pricing constraints we could not afford multiple random‑seed repetitions or large hyper‑parameter search. In future work, we will investigate lower‑cost surrogate generators or synthetic pre‑training so that broader statistical validation can be carried out without prohibitive expense.

\bibliography{aaai25}

\appendix

\end{document}